\documentclass[10pt,twocolumn,letterpaper]{article}

\usepackage{iccv}
\usepackage{times}
\usepackage{epsfig}
\usepackage{graphicx}
\usepackage{amsmath}
\usepackage{amssymb}

\usepackage{enumitem}
\usepackage{graphicx}
\usepackage{url}

\usepackage{color}
\usepackage{amsmath,amsfonts,amssymb}
\usepackage{graphicx}

\DeclareMathOperator{\E}{\mathop{\mathbb{E}}}

\newcommand{\Expect}{{\rm I\kern-.3em E}}

\newcommand{\KL}{\text{KL}}

\newcommand{\bI}{\boldsymbol{I}}

\newcommand{\bpsi}{\boldsymbol{\psi}}

\newcommand{\bmu}{\boldsymbol{\mu}}

\newcommand{\bSigma}{\boldsymbol{\Sigma}}

\newcommand{\btheta}{\boldsymbol{\theta}}

\newcommand{\bepsilon}{\boldsymbol{\epsilon}}

\newcommand{\by}{\boldsymbol{y}}
\newcommand{\br}{\boldsymbol{r}}

\newcommand{\bc}{\boldsymbol{c}}
\newcommand{\bm}{\boldsymbol{m}}
\newcommand{\bx}{\boldsymbol{x}}

\newcommand{\bW}{\boldsymbol{W}}
\newcommand{\bF}{\boldsymbol{F}}

\newcommand{\obs}{\mbox{\tiny$\mathcal{O}$}}

\newcommand{\obsi}{\mbox{\tiny$\mathcal{O}_i$}}

\usepackage[pagebackref=true,breaklinks=true,letterpaper=true,colorlinks,bookmarks=false]{hyperref}

\iccvfinalcopy 

\def\iccvPaperID{****} 

\ificcvfinal\fi
\begin{document}

\title{Unsupervised Data Imputation via \\Variational Inference of Deep Subspaces}

\author{Adrian V. Dalca\\
	MIT and MGH\\
	{\tt\small adalca@mit.edu}
	\and
	John Guttag\\
	MIT\\
	{\tt\small guttag@mit.edu}
	\and
	Mert R. Sabuncu\\
	Cornell University\\
	{\tt\small msabuncu@cornell.edu}
}


\maketitle

\begin{abstract}
	A wide range of systems exhibit high dimensional incomplete data. Accurate estimation of the missing data is often desired, and is crucial for many downstream analyses. Many state-of-the-art recovery methods involve supervised learning using datasets containing full observations. In contrast, we focus on unsupervised estimation of missing image data, where no full observations are available - a common situation in practice. Unsupervised imputation methods for images often employ a simple linear subspace to capture correlations between data dimensions, omitting more complex relationships. In this work, we introduce a general probabilistic model that describes sparse high dimensional imaging data as being generated by a deep non-linear embedding. We derive a learning  algorithm using a variational approximation based on convolutional neural networks and discuss its relationship to linear imputation models, the variational auto encoder, and deep image priors. We introduce sparsity-aware network building blocks that explicitly model observed and missing data. 
	We analyze proposed sparsity-aware network building blocks, evaluate our method on public domain imaging datasets, and conclude by showing that our method enables imputation in an important real-world problem involving medical images. The code is freely available as part of the \verb|neuron| library at~\url{http://github.com/adalca/neuron}.
\end{abstract}

\section{Introduction}


Highly incomplete data are found in a wide variety of domains. Sensor failure, occlusions, or sparsity by design all lead to missing data. For example, LIDAR scan data, providing depth measurements in a variety of problems, yield sparse point clouds~\cite{chen2017multi,li2016vehicle,uhrig2017sparsity}. 
Our work is motivated by a challenging real world medical imaging problem. In many clinical settings, medical image scanning time is limited by cost and physical or patient care constraints, leading to severely under-sampled images~\cite{dalca2019medical,dalca2017population,rousseau2010non}. For example, in many clinical settings, only every sixth 2D slice is acquired in a 3D MRI scan, resulting in 83\% of the anatomical data being missing. Estimating the missing data can yield meaningful clinical insight and help with downstream tasks such as registration and segmentation. 

State of the art methods for imputation, or estimation of missing values, often use statistics learned across the entire dataset. \textit{Supervised} methods that fill in missing values rely on datasets of full observations to learn relationships between present and missing data~\cite{bertalmio2000image,dong2016image,freeman2002example,kim2016accurate,xie2012image,yeh2016semantic}. For example,  super-resolution methods which can be viewed as imputation of higher-resolution pixel data require high resolution data to learn image statistics~\cite{dong2016image,freeman2002example,kim2016accurate}. In addition, super-resolution methods operate on a regular grid which is not applicable in many missing data settings. Image inpainting methods exploit statistical structure learned from images to impute a missing region, but often require densely observed regions of the existing images to learn image statistics~\cite{bertalmio2000image,xie2012image,yeh2016semantic}. Other subspace methods require fully observed data to learn a low-dimensional data representation, and then use these representations to impute missing information in sparsely observed data. 

\begin{figure}[tb!]
	\begin{center}
		\includegraphics[width=0.89\linewidth]{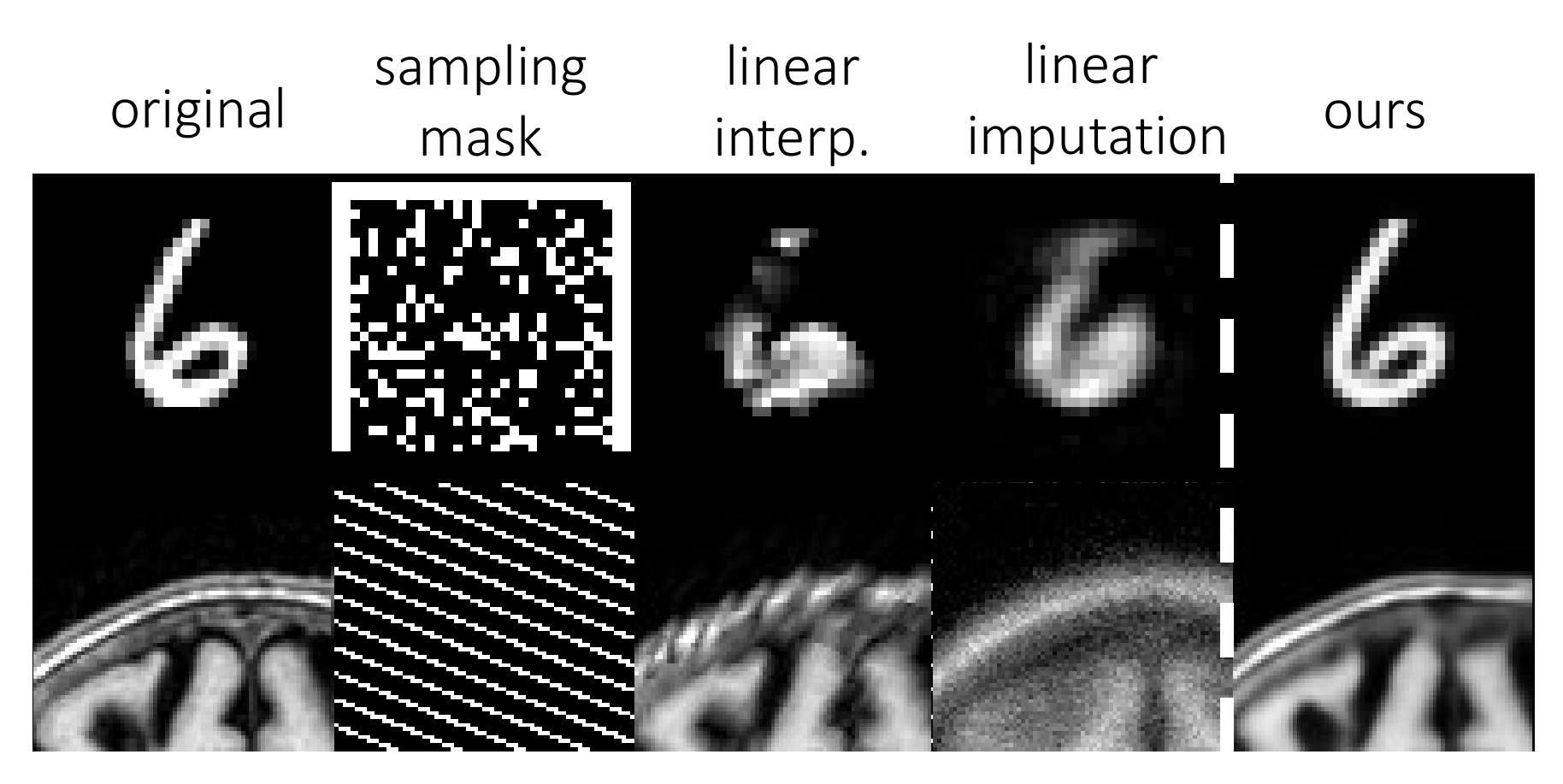}
	\end{center}
	%
	\caption{\textbf{Preview of unsupervised imputation for MNIST and (crop of) brain MRI slice.} We leverage a collection of images with missing pixels but common structure to successfully impute missing data in the presence of extreme sparsity. 
	}
	\label{fig:teaser}
\end{figure}

However, in many problems, fully observed data is unavailable or difficult to acquire. In this work, we focus on the recovery of missing image data in an \textit{unsupervised} setting where no full observations are available, but some common structure exists across the dataset. 
Unsupervised statistical imputation methods, spanning the literature from dictionary learning, factor analysis, manifold learning, and Principal Component Analysis (PCA) and its variants, often use linear subspace models to capture covariation across a dataset~\cite{bishop1995neural,little2014statistical}. These models assume each observed data point is a noisy, sparse observation of a lower dimensional linear subspace. They have been succesfully applied in areas such as network traffic flows~\cite{lakhina2004diagnosing} or collaborative filtering~\cite{sarwar2001item}, where the high dimensional data can often be well represented by a linear embedding. However, in many settings such as imaging, linear models are insufficient in capturing data representations~\cite{lee2009convolutional,oquab2014learning,radford2015unsupervised}. 



In this work, we propose a probabilistic model that frames high dimensional imaging data with very sparse pixel observations as being generated from a \textit{non-linear} subspace. Starting from this model, we derive a variational learning strategy that employs developments in deep neural networks and variational inference. We introduce building blocks of sparsity-aware deep architectures, and present a data inference method, based on this learning model, that enables both conditional mean imputation and multiple imputation of missing data. We discuss connections and important differences with linear imputation methods, which can be viewed as linear instantiations of our model, with the variational auto encoder, and with deep image priors. We evaluate this approach on unsupervised imputation in several public datasets comparing several model variants and linear alternatives. We analyze individual building blocks, and show that our method enables imputation of medical images in a real-world problem. Finally, we provide a comparative analysis with deep image priors.

\section{Related Works}


Collaborative filtering systems include only a sparse observation of users preferences~\cite{linden2003amazon,sarwar2001item}. Here, methods aim to learn from user preference to produce future recommendations. Often, these models build user representations using matrix completion methods, which share a goal with data imputation using linear embeddings. Recent methods have exploited convolutional neural networks for joint user representations with external information regarding content~\cite{li2017collaborative}. Other methods use shallow auto-encoders with sparse data and propose specific regularized loss functions~\cite{salakhutdinov2007restricted,sedhain2015autorec}. Similar to linear subspace models, these methods can be characterized as instantiations of our model.


Variational Bayes auto-encoders (VAEs) and similar models have been used to learn probabilistic generative models, often in the context of images~\cite{kingma2014semi,kingma2013auto,ranganath2014black}. Similarly, deep denoising auto-encoders use neural networks to obtain embeddings that are robust to noise~\cite{vincent2010stacked}. Our method builds on these recent developments to approximate subspaces using neural networks. Importantly, we show that a principled treatment of a sparsity model leads to important and intuitive differences from the VAE.

Deep Image Priors (DIP) use a generative neural network as a structural prior, and can be used to synthesize missing data~\cite{ulyanov2017deep}. For each image independently, the method finds network parameters that best explain the observed pixels. However, as parameters are image specific, this method is not amenable to extreme sparsity where image structure is hard to learn from the few observations of that image. Below, we discuss how our method is similar to DIP, how it differs, and perform a comparison in our experiments.

Several methods define \textit{sparse} neural networks in other contexts that are not directly related to our task, but still share nomenclature. For example, spatially-sparse CNNs assume a fully observed input, but the content itself is sparse, such as thin writing on a black background~\cite{graham2015,graham2017}. Faster sparse convolutions are proposed by explicitly operating on the pixels that represent content, with the focus of efficient computation. Other methods propose sparsity of the parameter space in neural networks to improve various metrics of network efficiency~\cite{han2015learning,liu2015sparse,wen2016learning}.

During the development of this work, several contemporaneous works have been shown to tackle related problems. Partial convolutions~\cite{liu2018image} have been developed to tackle image inpainting, where parts of the desired images are fully observed. A recent method uses adversarial training to guide a generator network to impute missing data, and introduces a discriminator \textit{hint} mechanism to enable training~\cite{yoon2018gain}. Within the medical image analysis domain, a recent method takes advantage of the similarity of local structure between different acquisition directions to enable subject-specific supervised training and imputation~\cite{zhao2018deep}. Outside of imaging-specific methods, recent papers have also shown similar development based on deep generative models in for imputing tabular and time series data~\cite{che2018recurrent,nazabal2018handling}.

\section{Methods}

\begin{figure*}[!bt]
	\begin{center}
		\includegraphics[width=1.0\textwidth]{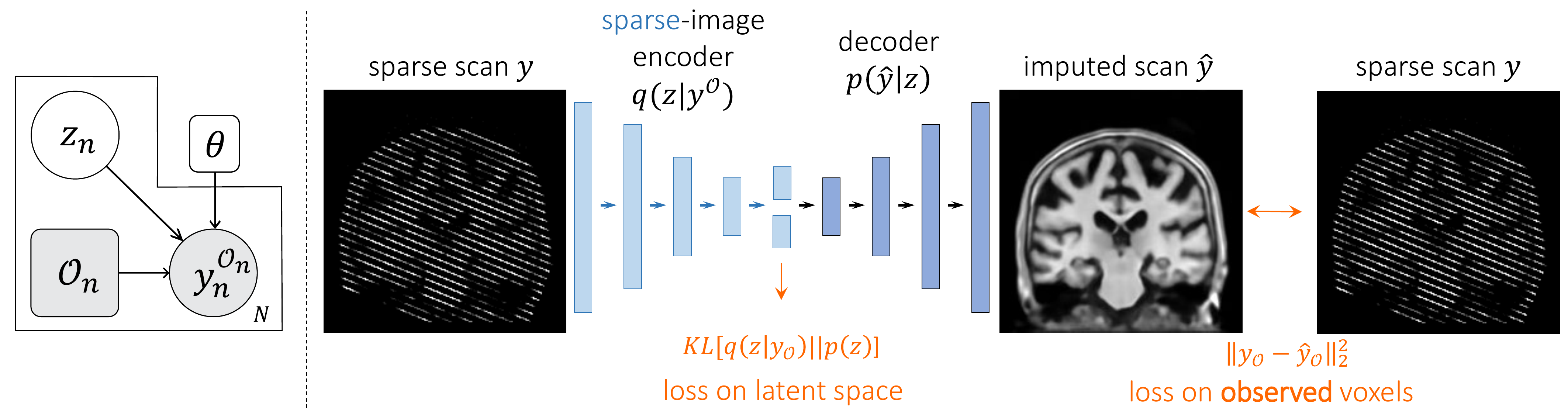}
	\end{center}
	\caption{\textbf{Overview} Left: graphical representation for our probabilistic model. Right: resulting neural network schematic, highlighting the encoder using \textit{sparsity-aware} layers, imputation via the decoder, and the \textit{sparsity-aware} loss function.}
	\label{fig:overview}
\end{figure*}

In this section we first present a general generative probabilistic model for sparse data observations using a non-linear subspace, and describe the imputation procedure using this model. We then show a learning algorithm that employs a variational approximation using neural networks, and introduce sparsity-aware neural network building blocks that explicitly model observed and missing data. Figure~\ref{fig:overview} illustrates an overview of the method.

\subsection{Model}

We let~$\by$ denote an image written as a vector of size~$D$ which we model as a high dimensional manifestation of a low-dimensional representation~$\bx$ of length~$d \le D$:
\begin{align}
\by &= f_{\btheta}(\bx) + \bepsilon, \quad \mbox{where} \quad \bepsilon \sim \mathcal{N}(0, \sigma^2 \bI_{D}),
\label{eq:main_model}
\end{align}
where~$\mathcal{N}(\bmu, \bSigma)$ denotes the multivariate normal distribution with mean~$\bmu$ and covariance~$\bSigma$,~$\bI_D$ denotes the $D \times D$ identity matrix, and~$f_{\btheta}(\cdot)$ is a (potentially non-linear) function parametrized by~${\btheta}$ that maps data from the low dimensional subspace to a full observation. The variance parameter~$\sigma^2$ captures independent pixel noise. We adopt a Gaussian prior for the low dimensional representation:
\begin{align}
\bx &\sim \mathcal{N}(0, \bI_{d}). 
\label{eq:prior}
\end{align}

We let~$\mathcal{O}$ indicate the set of observed entries in data~$\by$, and~$\by_{\obs}$ the corresponding vector of observed values. The set~$\mathcal{O}$ varies for each datapoint and is assumed to be small (representing high sparsity). The likelihood of an entire observed dataset~\mbox{$\mathcal{Y}_{\obs} = \{\by_{i,\obsi}\}$}, where~$\by_{i,\obsi}$ are observed data points, is therefore:
\begin{align}
p(\mathcal{Y}_{\obs}; \sigma^2)  
&= \prod_i \int_{\bx_i} p_{\btheta}(\by_{i,\obsi}|\bx_i) p(\bx_i) d{\bx_i}  \nonumber \\
&= \prod_i \int_{\bx_i} \mathcal{N}(\by_{i,\obsi} ; f_{\btheta}(\bx_i), \sigma^2\bI_D) \mathcal{N}(\bx_i; \boldsymbol{0}, \bI_d) d{\bx_i}. \nonumber
\label{eq:local-gmmppca-likelihood}
\end{align}

\subsection{Imputation}

We aim to infer the full data point~$\by$ given a \textit{sparse} observation~$\by_{\obs}$ using the posterior~$p({\by} | \by_{\obs})$:
\begin{align}
\log p_{\btheta}({\by} | \by_{\obs}) &= \log \int_{\bx} p_{\btheta}(\by|\bx) p_{\btheta}(\bx|\by_{\obs}) d\bx \nonumber\\
&=  \log \E_{p_{\btheta}(\bx|\by_{\obs})}[p_{\btheta}(\by|\bx)] \nonumber\\
&\overset{(*)}{\ge}   \E_{p_{\btheta}(\bx|\by_{\obs})}[\log p_{\btheta}(\by|\bx)] \\
&=  -\frac{1}{\sigma^2} \E_{p_{\btheta}(\bx|\by_{\obs})}[(\by - f_{\btheta}(\bx))^2] + \mbox{const}. \nonumber
\end{align}
where we used Jensen's inequality in ~$(*)$ and model~\eqref{eq:main_model} in the last line. To impute data, we use maximum-a-posteriori (MAP) estimation, and approximate it via the lower bound
\begin{align}
\hat{\by} =  \E_{p_{\btheta}(\bx|\by_{\obs})}[f_\theta(\bx)] < \arg\max_{\by} \log p_{\btheta}(\by | \by_{\obs}).
\label{eq:ymax}
\end{align}
%


Statistical imputation of the form~\eqref{eq:ymax} is referred to as \textit{conditional mean} imputation, which can underestimate the variance of downstream tasks~\cite{little2014statistical}. It is sometimes desirable to be able to sample the posterior itself thus producing \textit{multiple} imputations that give an indication of the variance captured by the model. Our method enables this process by sampling from the posterior approximation~\mbox{$\bx_k \sim p_{\theta}(\bx|\by_{\obs})$}, followed by 
\begin{align}
\hat{\by} \sim \mathcal{N}(f_\theta(\bx_k), \sigma^2\bI_D).
\end{align}
This process projects a sparse data point~$\by_{\obs}$ to a plausible representation~$\bx_k$, and then estimates a possible data point~$\hat{\by}$.

\subsection{Learning}

Unfortunately, computing the expectation~\eqref{eq:ymax} or sampling from~$p_{\theta}(\bx|\by_{\obs})$ is intractable.  Building on recent methods in variational inference such as the VAE, we employ an approximate variational posterior probability~$q_{\bpsi}(\bx|\by_{\obs}) \approx p(\bx|\by_{\obs})$ parametrized by~$\bpsi$, and minimize the KL divergence with the true posterior~\cite{jaakkola2000bayesian,jordan1999introduction,kingma2014semi,kingma2013auto}:
\begin{align}
&\min_{\psi} \KL \left[q_{\bpsi}(\bx|\by_{\obs}) || p(\bx|\by_{\obs})  \right] \nonumber \\
&= \min_{\psi} \E_{q} \left[ \log q_{\bpsi}(\bx|\by_{\obs}) - \log p_{\theta}(\bx, \by_{\obs}) \right] + \log p_{\btheta}(\by_{\obs}) \nonumber \\
&= \min_{\psi} \KL \left[  q_{\bpsi}(\bx|\by_{\obs}) ||  p(\bx)  \right] - \E_{q} \left[ \log p(\by_{\obs} | \bx ) \right],
\label{eq:VLB}
\end{align}
which is the negative of the \textit{variational lower bound} of the model evidence~\cite{kingma2013auto}.
We model the approximate posterior as a multivariate normal:
\begin{align}
q_{\bpsi}(\bx | \by_{\obs} ) = \mathcal{N}(\bx ; \bmu_{x | y_{\obs}}(\by_{\obs}), \bSigma_{x | y_{\obs}}(\by_{\obs})),
\end{align}
where~$\bSigma_{x | y}$ is diagonal. 
This approximation enables efficient sampling, facilitating imputation using
\begin{align}
\hat{\by} = \E_{q_{\psi}(\bx|\by_{\obs})}[f_\theta(\bx)] \simeq \frac{1}{K} \sum_k f(\bx_k)
\end{align}
where~$\bx_k$ are samples from~$\bx_k \sim q_{\psi}(\bx|\by_{\obs})$.

\begin{figure}[!bt]
	\begin{center}
		\includegraphics[width=0.5\textwidth]{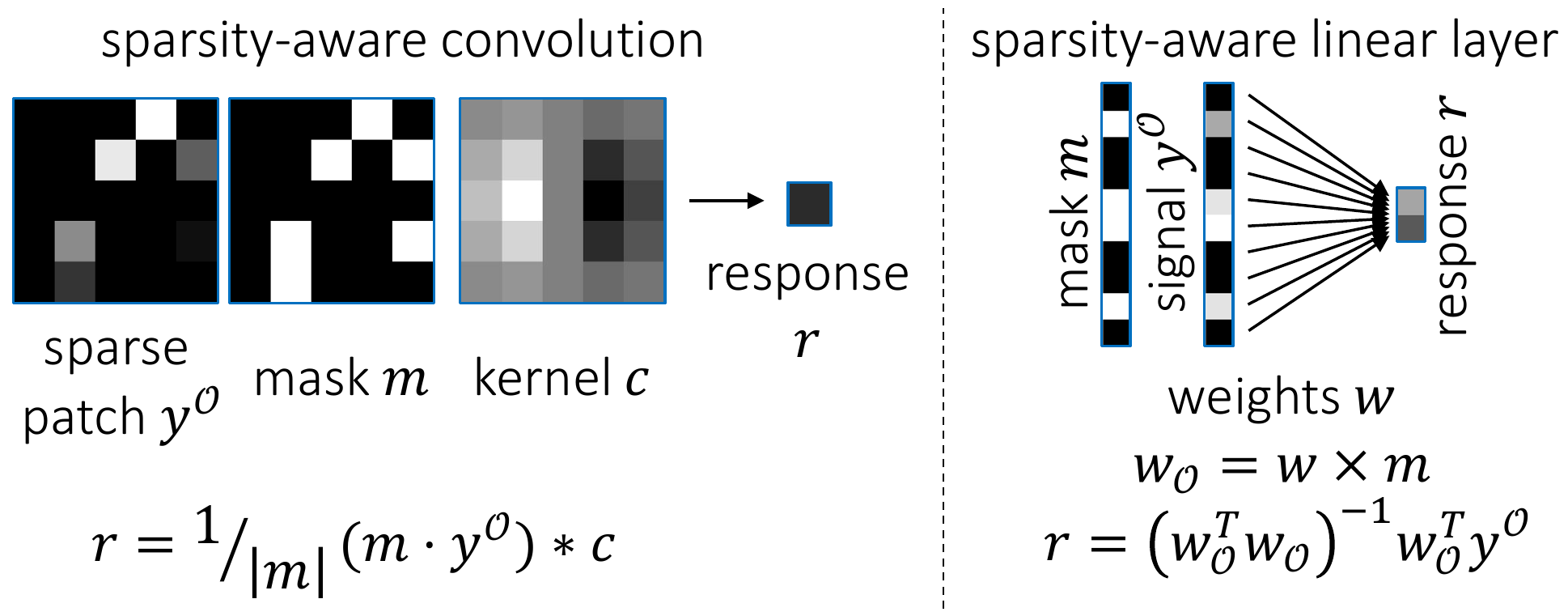}
	\end{center}
	\caption{Schematic examples of sparsity-aware convolution and linear layers, using the sparsity masks to correct conventional layer equivalents.}
	\label{fig:sparse-ops}
\end{figure}

\subsection{Learning via sparsity aware neural networks}
\label{sec:learning}

The functions~$(\bmu_{x | y_{\obs}}(\cdot), \bSigma_{x | y_{\obs}}(\cdot))$ take only the observed entries of~$\by$ and compute the subspace statistics. We estimate these using a neural network~$\text{enc}_{\bpsi}(\by_{\obs})$, parameterized by~$\bpsi$. We similarly approximate the generating function~$f_{\btheta}(\cdot)$ by a neural network~$\text{dec}_{\btheta}(\bx)$, parameterized by~$\btheta$. We jointly learn the parameters~$\{\bpsi, \btheta\}$ by optimizing the variational lower bound~\eqref{eq:VLB} using stochastic gradient methods. 
Specifically, for each data point~$\by_{i,\obs_i}$ with observed data~$\obs_i$, the resulting loss is:
\begin{align}
&\mathcal{L}(\bpsi, \btheta; \by_{i,\obs_i}) \nonumber\\
&= - \Expect_{\bx_k \sim q} \left[ \log p(\by_{i,\obsi} | \bx_k ) \right] + \KL \left[  q_{\bpsi}(\bx|\by_{i,\obsi}) ||  p(\bx)  \right]  \nonumber\\
&= \frac{1}{K\sigma^2} \sum_k ||\by_{i,\obsi} -  [f(\bx_k)]_{\obsi}||^2  \nonumber \\
&+  \text{tr}(\bSigma_{x|y_{\obs}} - \log|\bSigma_{x|y_{\obs}}|) + \bmu_{\bx | \by_{\obs} }^T \bmu_{\bx | \by_{\obs} } + \text{const}, \label{eq:main_loss}
\end{align}
where~$\bx_k$ are samples from~$q_{\psi}(\bx|\by_{i,\obs})$, and~$K$ is the number of samples we draw for each input image. The first term encourages the \textit{observed} entries of~$\by_i$ to be well recovered by the decoder. The second term encourages the subspace posterior~$q_{\bpsi}(\bx|\by_{\obs})$ to be close to the prior~$p(\bx)$. Since this posterior depends on only \textit{unobserved} entries, below we introduce several sparsity-aware building blocks for neural networks that handle sparse inputs (Figure~\ref{fig:sparse-ops}.

\paragraph{Fully Connected} Fully Connected (or dense) layers enable learning of broad correlations across pixels of a datapoint, and are extensively used in both imaging and non-imaging data.
In general, a fully connected layer can be written as $\br=\bF\by + \bmu_r$, where $\br$ is the output (response), $\by$ is the input, $\bF$ is the weight matrix and $\bmu_r$ is a bias term. When the input is partially observed as $\by_{\obs}$, a naive computation of the output might involve the columns of $\bF$ that correspond to the observed entries, $\bF_{\obs}$, via $\br=\bF_{\obs} \by_{\obs} + \bmu_{r, \obs}$. This is equivalent to filling in the missing $\by$ values with zeros. This approach, however, does not account for or exploit possible dependencies between entries of $\by$. We propose an alternative strategy based on linear models for imputation~\cite{little1989analysis}, where we adopt a linear model $\by=\bW \br + \bmu_y$. Given observed $\by_{\obs}$, the response $\br$ can be computed as:
\begin{align}
\br = (\bW_{\obs}^T\bW_{\obs})^{-1}\bW^T_{\obs}(\by_{\obs} - \bmu_{y,\obs}),
\end{align}
where $\bW_{\obs}$ are the rows of $\bW$ that correspond to the observed indices. We propose to use this formulation as a sparsity-aware fully connected layer, where~$\bW$ and~$\bmu_y$ are now the layer parameters.
%
We discuss the linear formulation to imputation, which motivated this layer, in the Subsection \textit{Comparison with Linear Subspace Models}. 
In our experiments, we demonstrate that this layer more accurately computes linear projections of sparse data and leads to improved training compared to a traditional fully connected layer.

\paragraph{Convolution} For imaging data, hierarchical convolutional operations extract meaningful representations. Existing methods often fill in missing values with a constant, such as zero or the mean value at that pixel across the dataset, but these do not usually accurately estimate the existing image signal~\cite{uhrig2017sparsity}.

Given a sparse input image, we experiment with two convolutional approaches. In the first, we apply a weighted convolution, where the convolution kernel~$\bc$ is modified to vary with image location~$k$ by the binary observed mask~$\bm$~\cite{uhrig2017sparsity}. The new filter response~$\br$ at location~$k$ is
\begin{align}
\br_k = \frac{|N(k)|}{\sum_j \bm_j}  \sum_{j\in N(k)} \bm_{k-j} \by_{k-j} \bc_{j}
\end{align}
where~$N(k)$ indicates all the pixels neighboring~$k$ within some filter kernel size. This weighted filter uses only the existing information in computing a response. We then compute the mask to be used at a subsequent layer~$\bm'$:
\begin{align}
\bm'_k = \Big( \sum_{j\in N(k)} \bm_{k-j} \Big) > 0.
\end{align}
Since convolutional layers can be applied hierarchically, even very sparse data will often lead to a dense deep feature response. This sort of convolution was recently used to impute LIDAR depth data in a supervised context~\cite{uhrig2017sparsity}.

Since we focus on image data, linear interpolation provides a rough approximation for missing pixels. Therefore, in a second strategy, we first approximate the missing data with linear interpolation, and provide the first convolution layer with both the observation mask and the interpolated image, as two input channels.

%


\subsubsection{Networks}

Different architecture families are appropriate for different problems and datatypes. We focus on architectural design decisions that pertain to utilizing missing data and account for the image content type, rather than describing specific details about particular architectures.

\begin{itemize}[leftmargin=1em]
	\item Hierarchical convolutional layers are used to extract image-space features. We experiment with (sparsity-aware) fully convolutional encoder and decoders.
	\item   In many domains, capturing covariation across a large image can provide useful structural information. We therefore explore encoders that use a (sparsity aware) fully connected layer following several convolutions, and a decoder that uses a fully connected layer followed by convolutions.
	\item For large data, such as volumetric medical scans, both convolutional and fully connected layers capture important relationships, but the volumes are too large to employ the designs above. In these cases, we propose an architecture that replaces the fully connected layer by \textit{locally} connected sparse layers, which affect separate subregions of the volume. 
\end{itemize}

\subsection{Connection to Other Models}
\label{sec:compare-discuss}

\subsubsection{Linear Subspace Methods}

We show that linear subspace methods are a specific case of our model. Let~$f(\bx) = \bW\bx + \bmu$, where weight matrix~$\bW$ controls the model covariance and~$\bmu$ is the data mean. This yields the sparse model 
\begin{align}
\by_{\obs} &= \bW_{\obs} \bx + \bmu_{\obs} + \bepsilon,
\label{eq:linear-obs-model}
\end{align}
where~$\bW_{\obs}$ selects the rows of~$\bW$ that correspond to observed entries in~$\by$. Using the proposed learning strategy~\eqref{eq:main_loss} and the pseudo-inverse of~$\bW$, we can choose the approximating posterior,
\begin{align}
q_{\bpsi}(\bx|\by_{\obs}) = \mathcal{N}(\bx; \bSigma_{\obs,\obs}\bW_{\obs}^T (\by_{\obs} - \bmu_{\obs}); \sigma^2 \bSigma_{\obs,\obs})
\end{align}
where~$\bSigma_{\obs,\obs}=(\bW_{\obs}^T\bW_{\obs})^{-1}$. Given learned parameters, we impute missing pixels using the expectation~\eqref{eq:ymax}
\begin{align}
&\arg\max_{\by} \E_{q_{\bpsi}(\bx|\by_{\obs})}[\bW\bx + \bmu] = \bW \E_{q_{\bpsi}(\bx|\by_{\obs})}[\bx] + \bmu \nonumber\\
&= \bW (\bW_{\obs}^T\bW_{\obs}  + \sigma^2\bI)^{-1}\bW_{\obs}^T  (\by_{\obs} - \bmu_{\obs}) + \bmu.
\label{eq:linear_imputation}
\end{align}
Intuitively, computing~$\E_{q_{\bpsi}(\bx|\by_{\obs})}[\bx]$ linearly projects the data point onto the low-dimensional subspace using only the \textit{observed} pixels. The recovered data point is then computed by linearly projecting the estimated low-dimensional representation into a high-dimensional space. The parameters~$\{\bW, \bmu, \sigma^2\}$ can be learned using the expectation maximization algorithm~\cite{little2014statistical}. Extended linear subspace models, such as ones that regularize the weights~$\bW$ or include mixtures of Gaussians, can similarly be seen as instantiations of our model. We used these concepts in proposing novel sparsity-aware fully connected (or dense) neural network layers above.

\begin{figure*}[tb!]
	\begin{center}
		\includegraphics[width=0.23\linewidth]{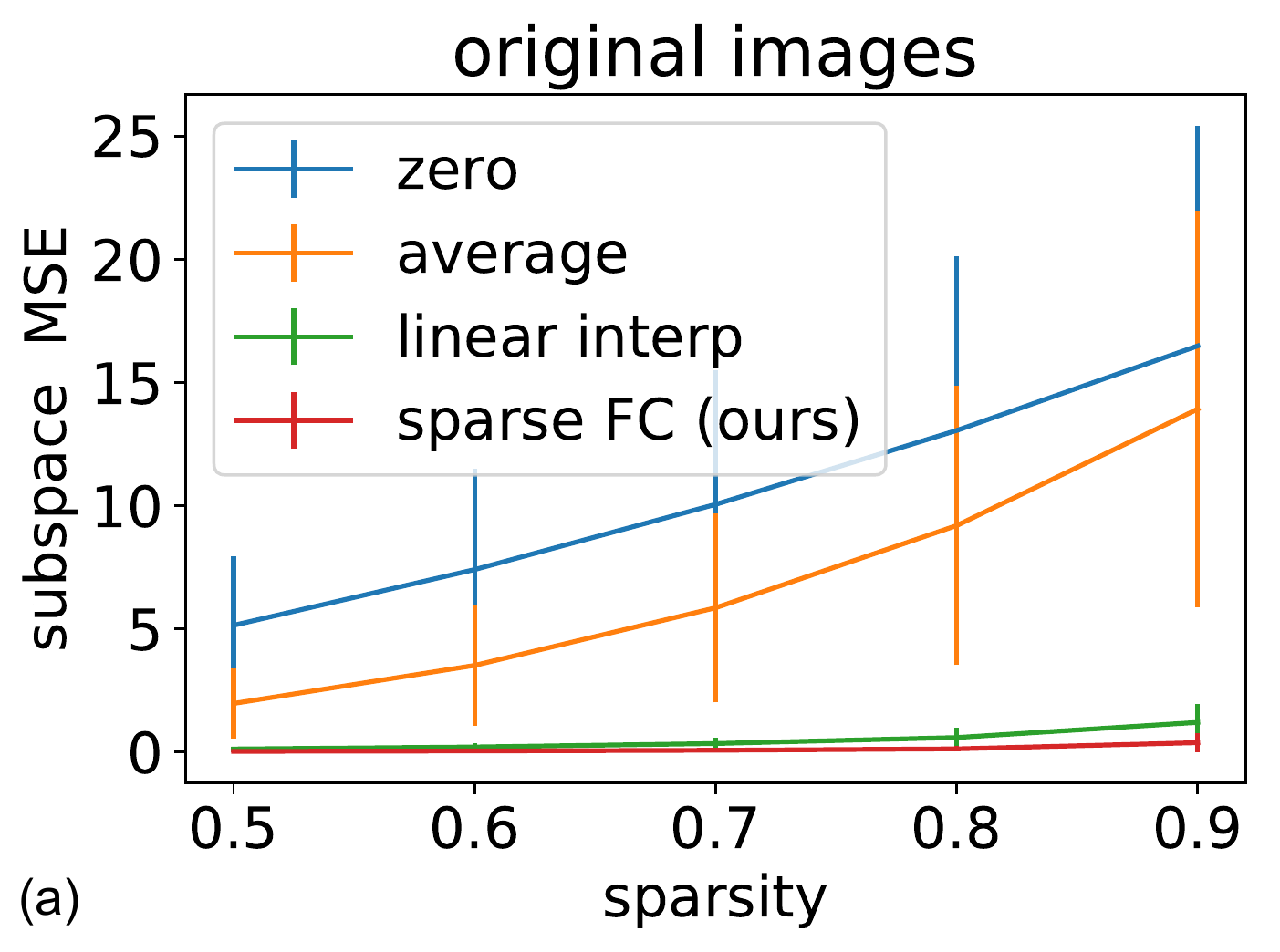}
		\includegraphics[width=0.23\linewidth]{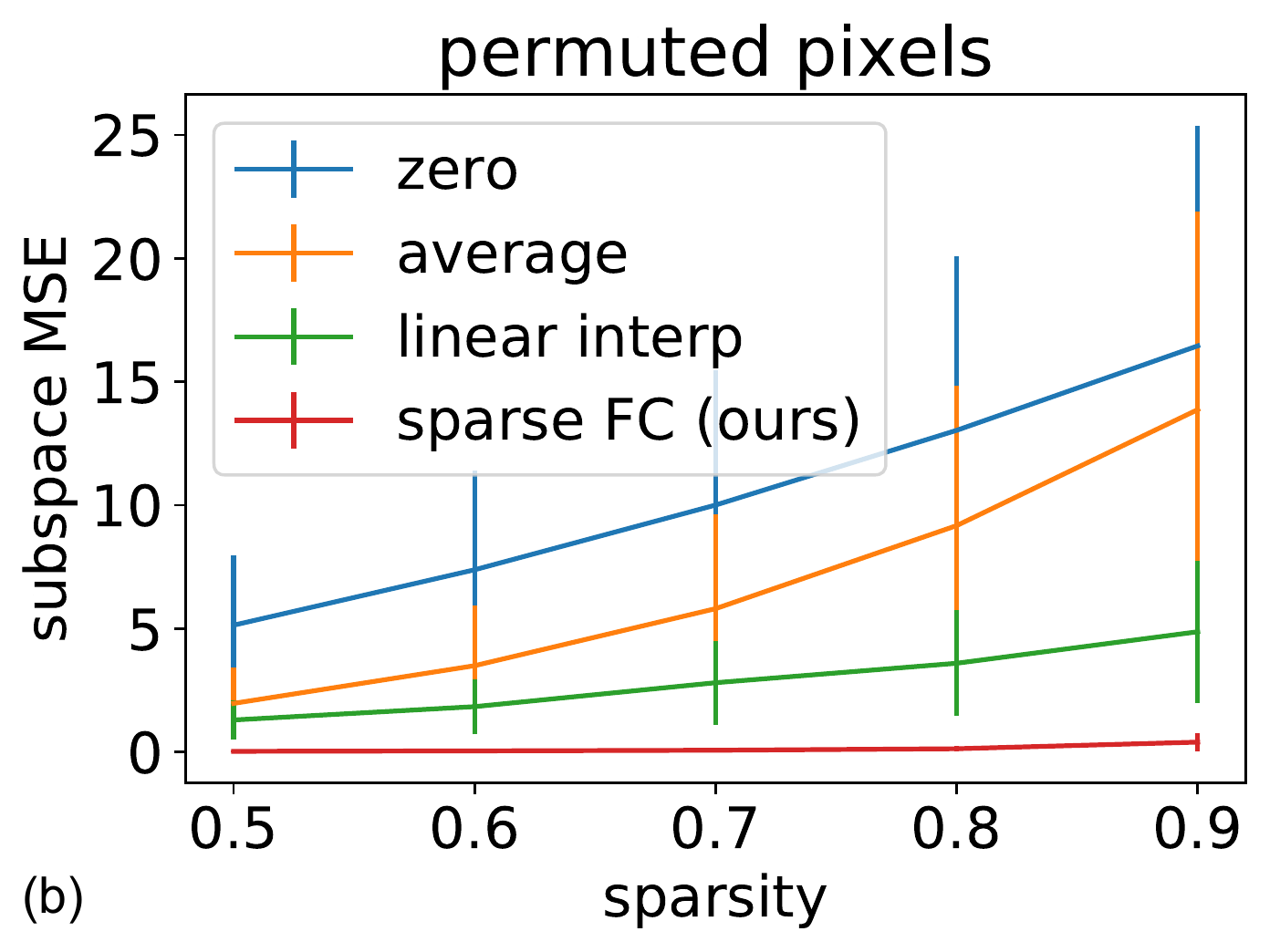}
		\includegraphics[width=0.23\linewidth]{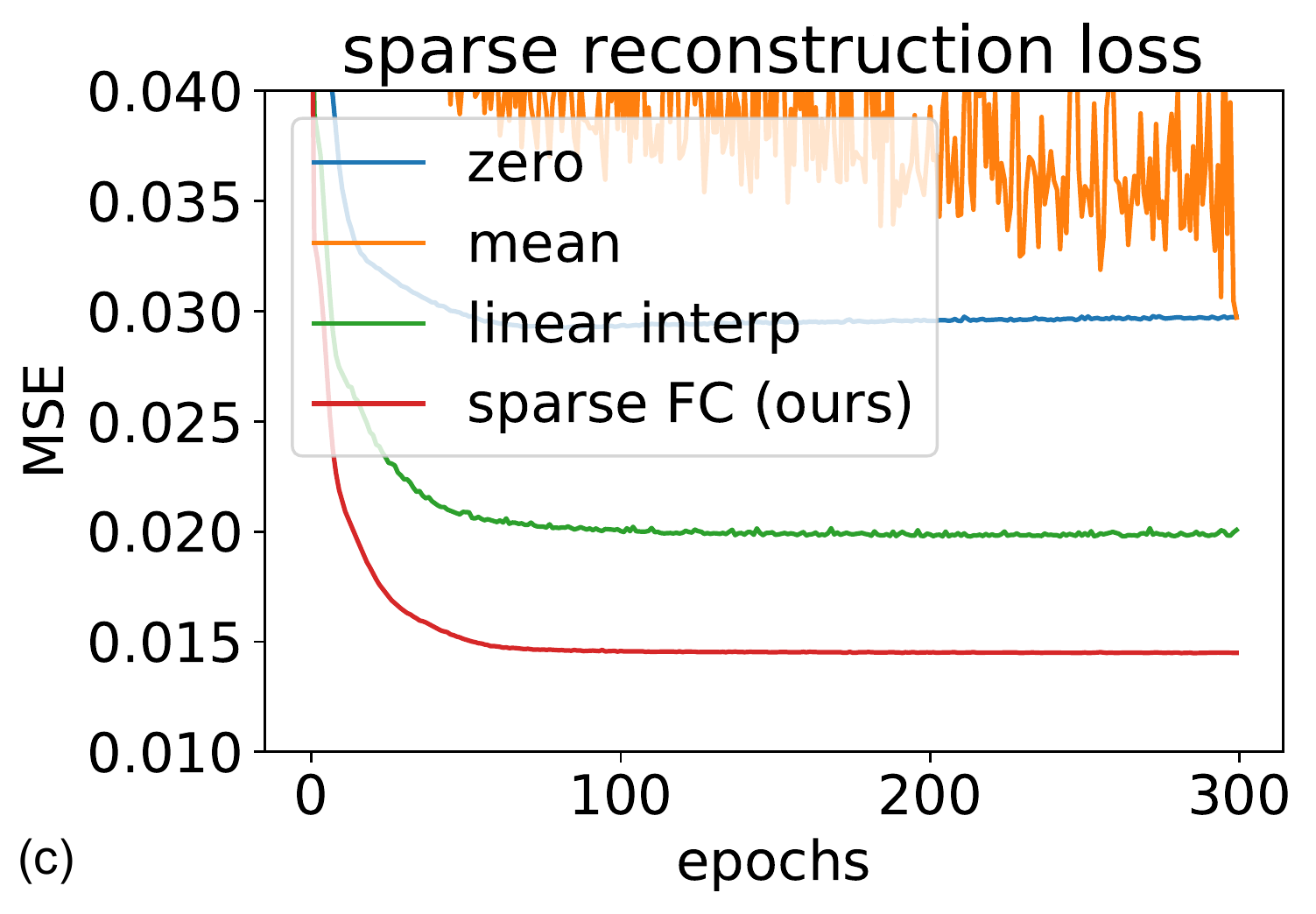}
		\includegraphics[width=0.24\linewidth]{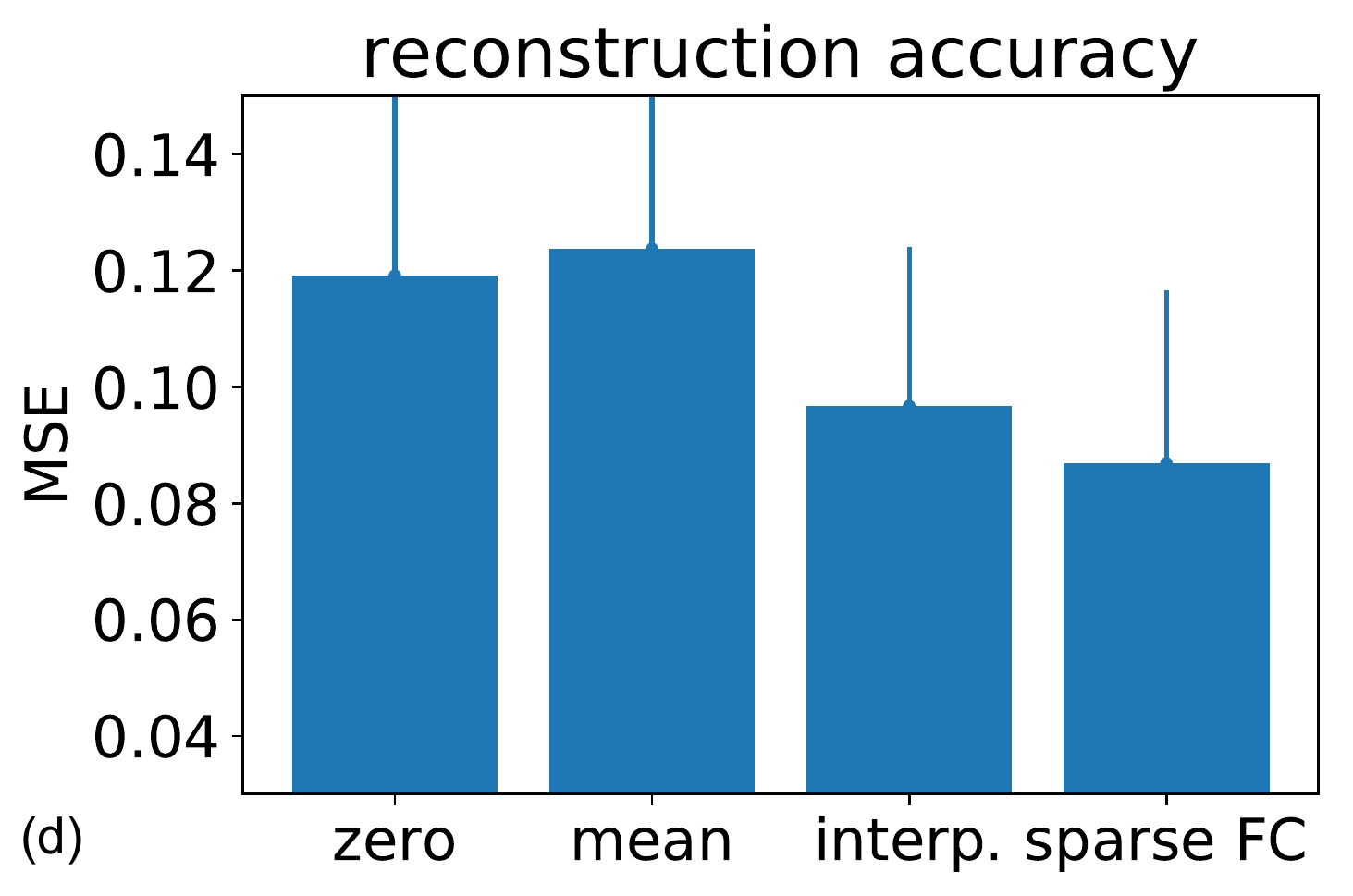}
	\end{center}
	%
	\caption{\textbf{Analysis of proposed sparse fully connected layer.} (a) and (b): subspace reconstruction error for three methods that use standard fully connected layer and our method, for the original images and images with permutted pixels. (c): linear auto-encoder convergence of sparse reconstruction loss on validation set. (d): linear auto-encoders full image reconstruction error. The proposed fully connected layer improves subspace estimation as well as the missing pixel values. 
	}
	\label{fig:fc_analysis}
\end{figure*}


\subsubsection{Variational Auto Encoder}

Our model and resulting learning strategy relate to the variational auto-encoder and other recent methods using approximations based on neural networks~\cite{kingma2014semi,kingma2013auto,ranganath2014black}.  A key difference is that our generative probabilistic model explicitly captures missing data. Our focus is on describing how this aspect changes generative models, providing a principled derivation of the learning strategy in the presence of missing data, and introducing missing data aware network layers. Specifically, in~\eqref{eq:main_loss} the posterior~$q_{\bpsi}(\bx|\by_{i,\obs})$ depends on only the \textit{observed} entries of~$\by_i$, leading to our network building blocks. The reconstruction term~$\Expect_{\bx_k \sim q} \left[ \log p(\by_{i,\obs} | \bx_k ) \right]$ is evaluated at only the observed voxels, enabling parameter learning in \textit{unsupervised} settings. In our experiments, we investigate how different parts of these methodological differences affect imputation.

\subsubsection{Deep Image Priors and Amortized Inference}

Deep Image Priors~\cite{ulyanov2017deep} use a generative neural network~$g_{\phi_{\by}}(\cdot)$ as a prior for image~$\by$, such that~\mbox{$\by=g_{\phi_{\by}}(\bx)$}. Parameters~$\phi_{\by}$ are optimized separately for each image. In the context of sparse data, this method can be used to synthesize missing pixels by first obtaining~\mbox{$\widehat{\phi}_{\by} = \arg\min_{\phi_{\by}} ||(g_{\phi_{\by}}(\bx))_{\obs} - \by_{\obs}||^2$} for some fixed~$\bx$, and then computing the full image~$\widehat{\by} = g_{\widehat{\phi}}(\bx)$. This strategy requires enough observed pixels in each image~$\by_{\obs}$ to be able to infer image structure and thus the missing data, and has been demonstrated in the case of large natural images with half of the pixels missing. 

In contrast, we focus on severe sparsity in each image, and leverage commonality across a dataset rather than the observed pixels in a single image. Our method learns a similar decoder network~$f_{\theta}(\bx)$ to be able to decode \textit{all} images in a dataset, rather than learning an image-specific generative network. In this sense, our decoding model can be seen as a collection-wide global version of the Deep Image Prior, where the embedding~$\bx$, estimated by the encoder~$p_{\theta}(\bx|\by_{\obs})$ specifies the instance to be recovered. 

In addition, our model can be seen as amortized inference over a collection of images with missing pixels. The general decoder learned by our model, together with an image-specific embedding~$\bx$, act as Deep Image Prior for the entire collection of images with common structure.

\subsection{Implementation}

We implement our method as part of the \verb|neuron| package~\cite{dalca2018priors}, which is available at~\mbox{\url{http://github.com/adalca/neuron}}, using Keras~\cite{chollet2015} with a Tensorflow~\cite{abadi2016} backend.


\section{Experiments}

We provide a series of experiments evaluating the presented method and its utility in the context of \textit{unsupervised} sparse datasets. 

We first analyze the proposed sparse fully connected layer to illustrate the improvement over traditional strategies in the sparse setting. We then evaluate several models on different image datasets to shed light on how various approaches perform in different settings. We focus on comparing image imputation using linear subspaces with several variants of our non-linear subspace model. Our goal is to demonstrate the potential of straightforward use of our method, rather than proposing an optimized architecture for a specific task. We then show the utility of our model on a clinical image imputation task. Finally, we analyze our algorithm compared to Deep Image Priors.

\begin{figure*}[t!]
	\begin{center}
		\includegraphics[width=0.77\linewidth]{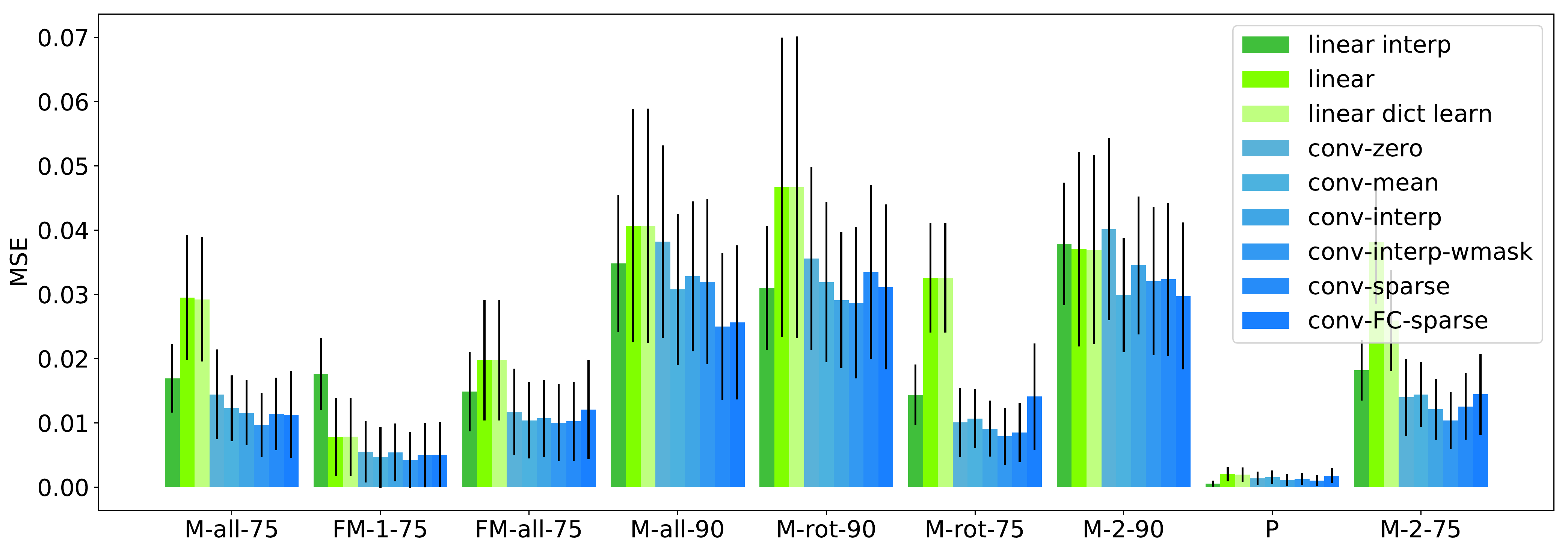}
		\includegraphics[width=0.19\linewidth]{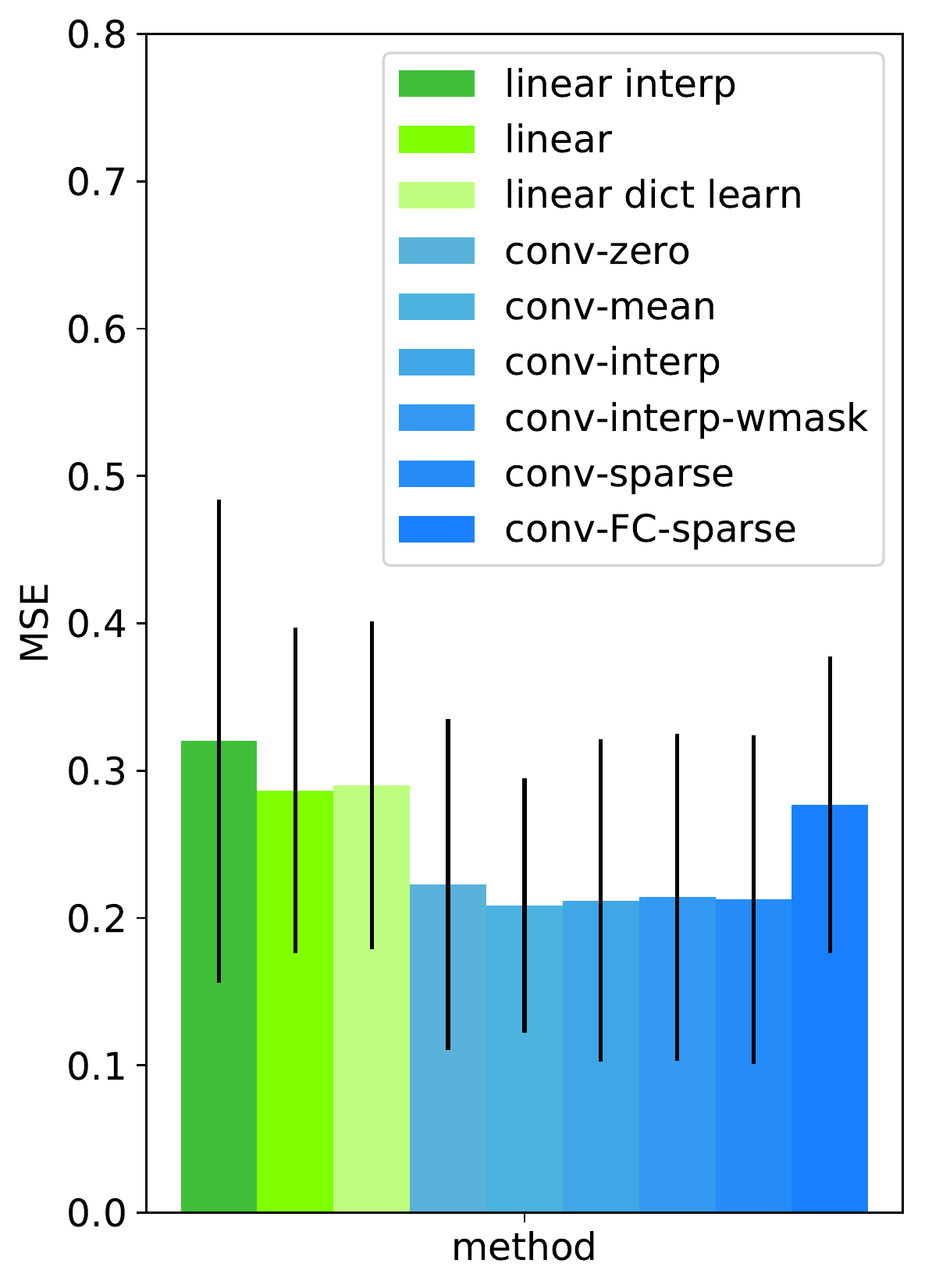}
	\end{center}
	%
	\caption{\textbf{Imputation result.} Left: Errors for reconstruction of the test ground truth for various datasets, using each imputation method. Datasets are shortened: MNIST to M, FASHION-MNIST to FM, and MRI Patches to P. The last number for each dataset name indicates the imposed sparsity (e.g. 75\% means 75\% percent of the pixels are missing). Right: Errors for reconstruction of the test ground truth for coronal MRI Slices. With the occasional exception of the simplistic \textit{conv-zero} strategy, our model variants consistently yield improved results compared to the baselines.
	}
	\label{fig:graph_results}
\end{figure*}

\subsection{Data}
We use three datasets in our experiments. First, the MNIST dataset consists of small (28x28 pixels) 2D images of hand-written digits~\cite{lecun1998mnist}. We create three variants: MNIST-2 which only consists of the digit 2, MNIST-all which refers to the original dataset, and MNIST-rot which contains all of the digits rotated at a random angle between 0 and 360 degrees. Second, we use the FASHION-MNIST dataset, which consists of 28x28 pixel images of 10 types of clothing items~\cite{xiao2017online}. These have more structure in each image compared to MNIST digits.  Finally, we use a large-scale, multi-site dataset of 7829 T1-weighted brain MRI scans, compiled from eight publicly available datasets: ADNI~\cite{mueller2005ways}, OASIS~\cite{marcus2007open}, ABIDE~\cite{di2014autism}, ADHD200~\cite{milham2012adhd}, MCIC~\cite{gollub2013mcic}, PPMI~\cite{marek2011parkinson}, HABS~\cite{dagley2015harvard}, and Harvard GSP~\cite{holmes2015brain}. Acquisition details, subject age ranges and health conditions are different for each dataset. We performed standard pre-processing steps on all scans, including resampling to~$1$mm isotropic voxels, and affine spatial normalization for each scan using FreeSurfer~\cite{fischl2012}. We crop the final images to $192\times 176 \times 224$.

In our experiments, we simulate the patterns of missing pixels, which we then remove from the images. We split each dataset into 50\%, 30\%, and 20\% for train, validation, and test sets respectively, all of which are sparse. The test set in each dataset is only evaluated once. We highlight, however, that our focus is on evaluating the models on the \textit{unsupervised} task of imputing missing pixels. Below, 90\% sparsity means 90\% of the data is missing.

\begin{figure}[tb!]
	\includegraphics[width=\linewidth]{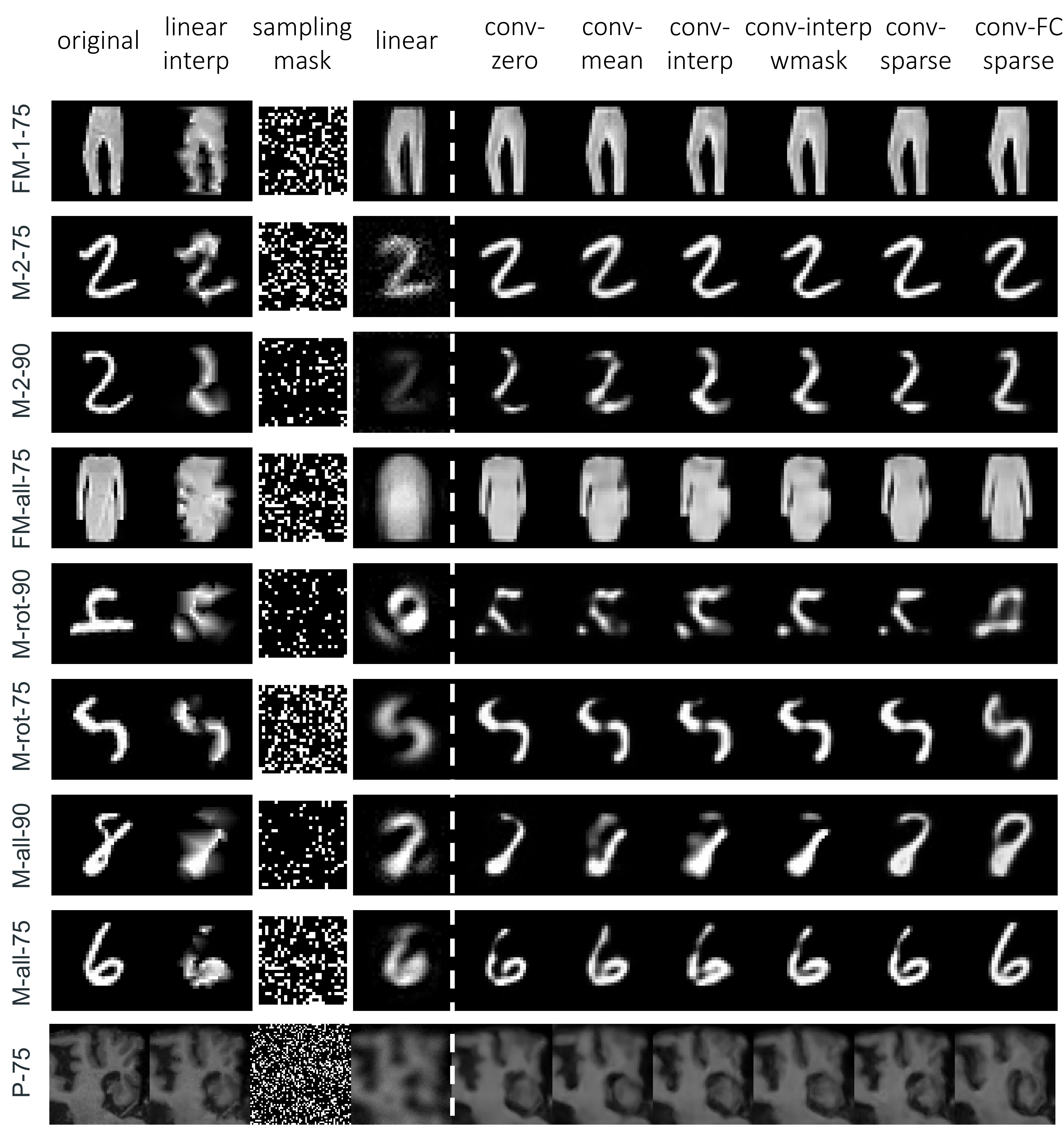}
	\caption{\textbf{Example Imputation results.} Each row is a separate dataset. The first three columns show the original ground truth image, the linear interpolation and the random sampling mask. The last seven columns illustrate a linear subspace model, and six variants of our method involving sparse fully connected layers. Other model variants show perceptually closer results to the true (unobserved) data.		
	}
	\label{fig:example_results}
\end{figure}

\subsection{Fully Connected Layer}

We first analyze the importance of the proposed sparse fully-connected layer using the FASHION-MNIST dataset (the other datasets result in comparable results). We compare several strategies for using fully connected layers with missing data, before activation and omitting the bias term. 

We compute the linear projection of fully-observed images onto a low-dimensional subspace of size~$10$ with weights initialized using PCA, and treat this as the ground truth projection. We then randomly sub-sample the data at several sparsity levels. We compute several baselines that use a conventional fully connected layer: filling the missing image pixels with 1) zeros, 2) the average value at each pixel location, or 3) linearly interpolated values.  Finally, we test our sparsity-aware fully connected layer. 

We compare the sparse image projections of each method with the ground truth response using mean squared error. We also consider a setting where spatial consistency is missing, often found in other domains, by randomly permuting image pixels and repeating the experiment. 

\begin{figure*}[tb!]
	\begin{center}
		\includegraphics[width=0.9\linewidth]{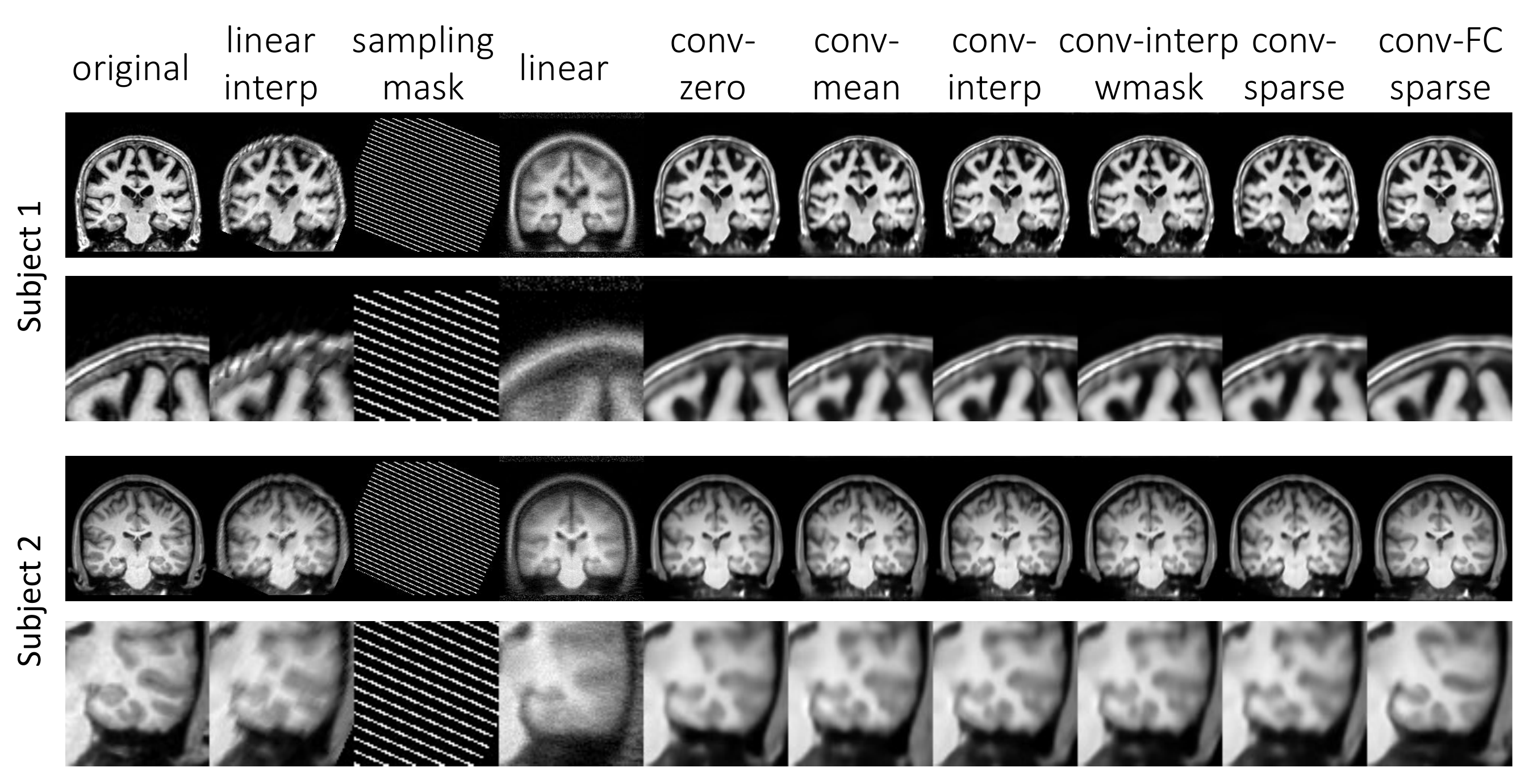}
	\end{center}
	%
	\caption{\textbf{Example of MRI slice imputation result.} For each subject scans, the top shows the overall slice, and the bottom zooms in on a portion of the image. The first three columns show the original ground truth image, the linear interpolation and the random sampling mask. The last seven columns illustrate a linear subspace model,  and six variants of our method involving sparse fully connected layers. All proposed models show perceptually similar results to the true (unobserved) data. 
	}
	\label{fig:mri_imputation}
\end{figure*}

Figure~\ref{fig:fc_analysis}(a,b) shows that using the sparsity-aware fully connected layer leads to a significantly improved projection followed by spatial interpolation. The random permutation dramatically affects the spatial interpolation baseline, but does not affect the other methods since they treat dimensions independently. 

In a second experiment, we learn a linear auto-encoder that uses the linear encoding methods described above, and a fully connected decoding layer. Figure~\ref{fig:fc_analysis}(c,d) reports the validation loss (on observed pixels only), and final validation reconstruction error (on all pixels). The results show that using sparsity-aware fully connected layers leads to improved reconstructions, promising to improve network performance in the context of sparse input data.


\subsection{Random Sparsity in Image Datasets}

In this section, we simulate random missing pixels for sparsity factors of 75\% and 90\% in images from MNIST-2, MNIST-all, MNIST-rot, and FASHION-MNIST. In addition, to evaluate more complex covariance structure in images, we obtain random 2D 64x64 patches from the brain MRI dataset. 

\subsubsection{Baselines}
In this section, our first baseline is the linear subspace model of~\eqref{eq:linear-obs-model} which is often used in sparse settings~\cite{little2014statistical}. We implement both stochastic gradient descent (SGD) based optimization, as well as Expectation Maximization ~\cite{little2014statistical}. For small datasets and models we find the two implementations to behave comparably, but the expectation maximization becomes intractable for large datasets and models. We therefore report results from the SGD implementation. In addition, we implement a (linear) sparse dictionary learning method using a L0-norm to encourage a sparse latent representation.

\subsubsection{Model Variants} 
We evaluate several variants of our method using sparsity-aware implementations, from simpler filling strategies to our full model:

\begin{itemize}[leftmargin=1em]
	\item \textit{conv-zero} fills missing image pixels with zeros and uses a normal convolutional encoder and the sparse loss~\eqref{eq:main_loss}
	\item \textit{conv-mean}  fills missing image pixels with the pixel-wise dataset mean and uses a normal convolutional encoder and the sparse loss
	\item \textit{conv-interp} linearly interpolates the missing pixels of an input image and uses a normal convolutional encoder and the sparse loss
	\item \textit{conv-interp-wmask} adds the incomplete data sampling mask as a model input channel to \textit{conv-interp}
	\item \textit{conv-sparse} uses a sparsity-aware, fully convolutional encoder and decoder architecture, each consisting of five residual blocks of two sparse convolutions with 3x3 kernels, ReLu activations and a 2x2 max-pooling or upsampling layer, and uses the sparse loss
	\item \textit{conv-FC-sparse} adds a sparsity-aware fully connected layer to the end of the encoder and the start of the decoder, enabling covariances across the entire image. We use an encoding of size 10 for MNIST and FASHION-MNIST, and 100 for the MRI patches.
\end{itemize}

\begin{figure*}[t!]
	\begin{center}
		\includegraphics[width=0.85\linewidth]{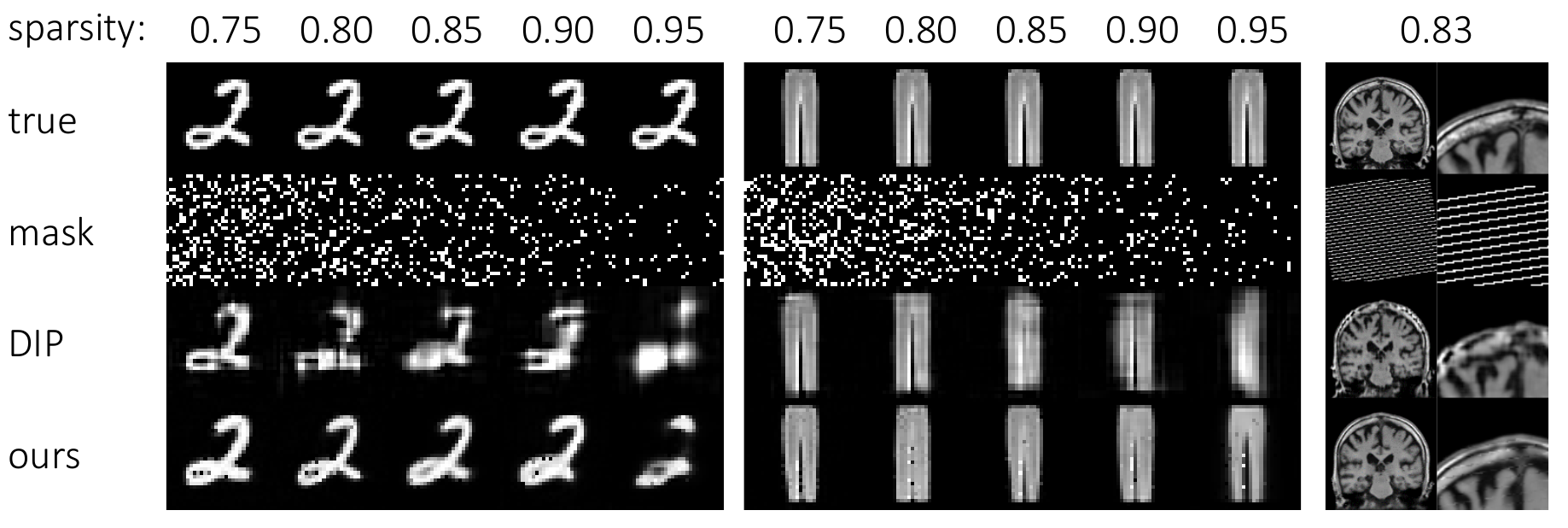}
		\includegraphics[width=0.35\linewidth]{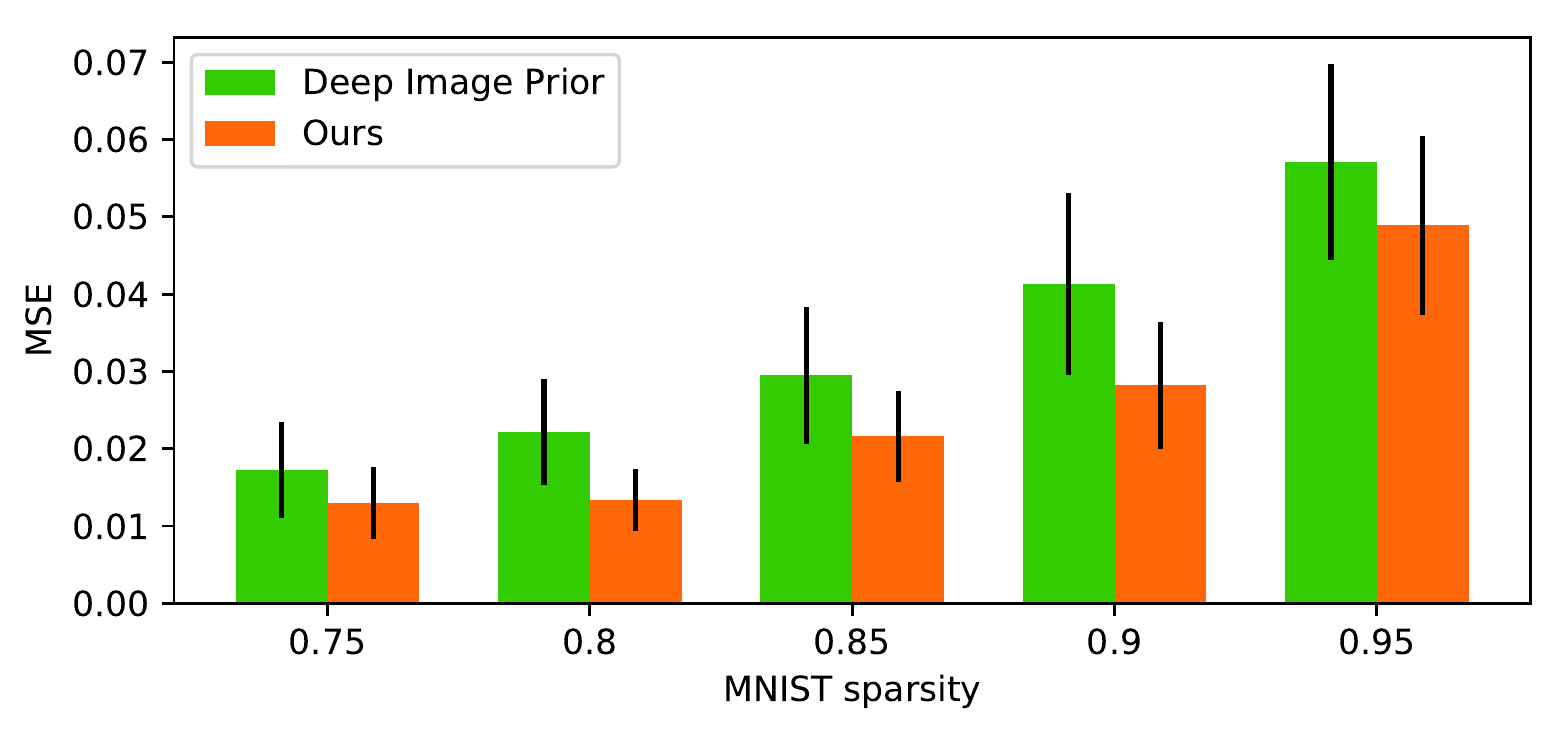}
		\includegraphics[width=0.35\linewidth]{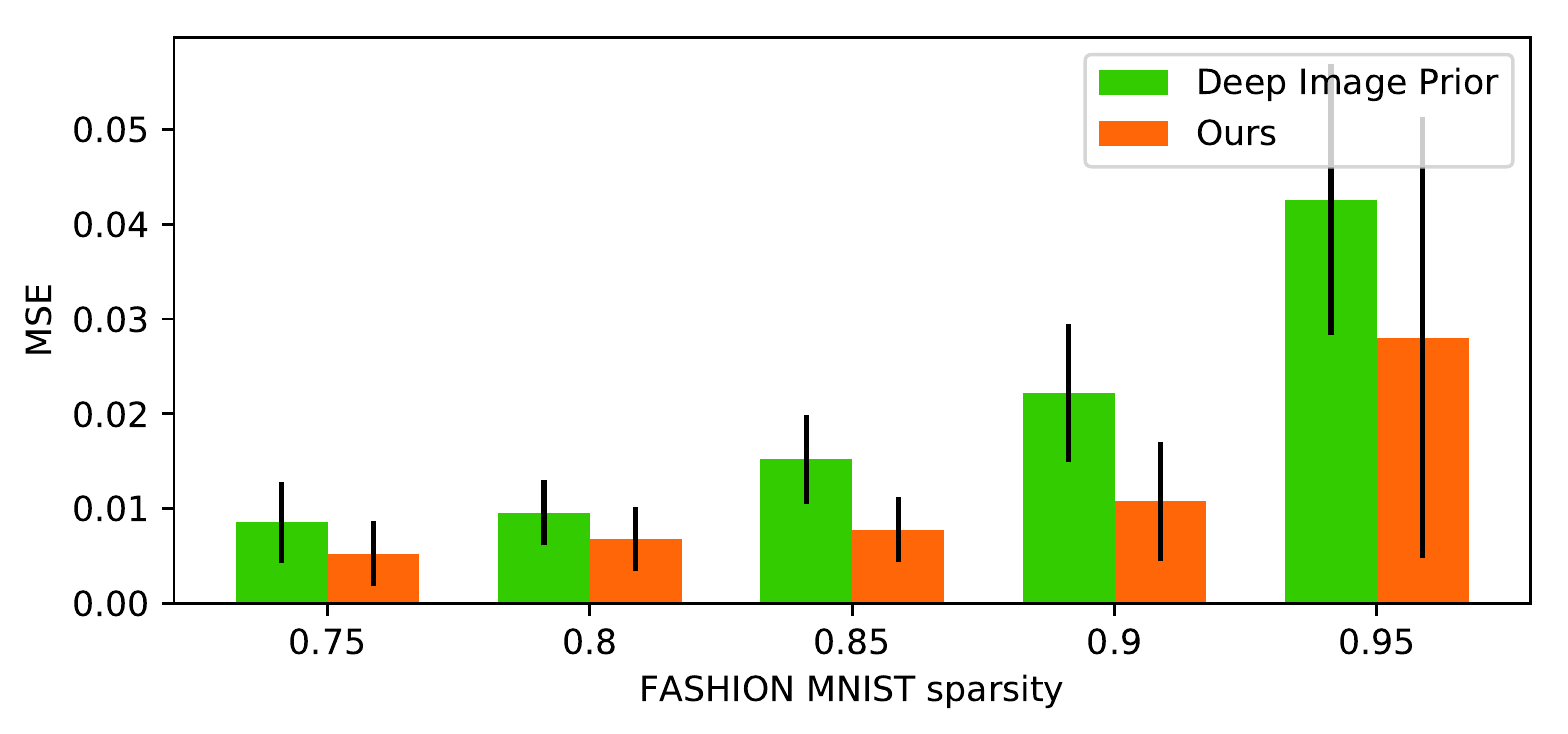}
		\includegraphics[width=0.17\linewidth]{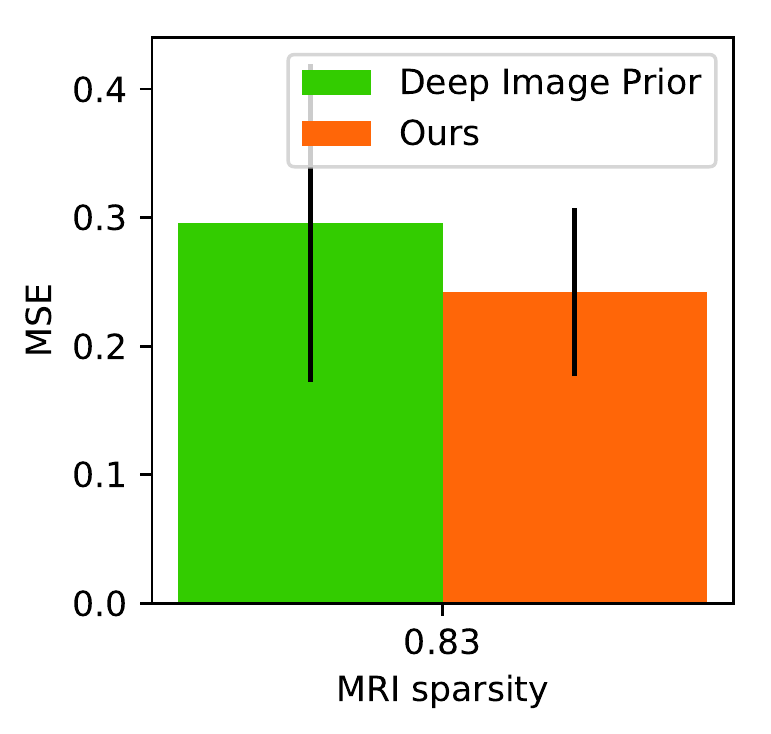}
	\end{center}
	%
	\caption{\textbf{Deep Image Prior Comparison.} \textbf{Comparison with Deep Image Priors} Top: Example instances for the MNIST-2, FASHION-MNIST-1, and sparse MRI datasets. The sparsity of the MRI images is determined by the slice spacing. Bottom: MSE plots for each of dataset. Our method (orange) has consistently better reconstruction by leveraging the entire dataset of images. 
	}
	\label{fig:dip}
\end{figure*}

\subsubsection{Results}
Figures~\ref{fig:graph_results} and~\ref{fig:example_results} illustrate the results. For the single digit, all methods, including the linear subspace are able to exploit covariation across the dataset, despite significant sparsity. However, as more variability is present in the full and rotated datasets, the linear subspace methods are unable to capture the image structure and properly impute the images. In contrast, non-linear subspace models are able to capture complex covariances, even in extreme sparsity situations. Using a sparsity-driven VAE after filling the missing pixels using linear-interpolation or mean-filling performs better than the linear models, and having explicit sparsity-aware convolutional encoder performs the best among all variants.

In general, several design decisions are possible for a given method, including varying encoding size, noise parameters and architecture designs. Here, our goal is to illustrate that in various datasets and sparsity levels, our method promises to dramatically improve imputation by employing a non-linear subspace.


\subsection{Brain MRI Slices}

We demonstrate the utility of our method on sparse brain MRI acquisition. In many clinical settings, scanning time is limited, leading to severely under-sampled medical scans. For example, in many clinical settings, only every sixth 2D slice is acquired in the 3D MRI scan. Moreover, because of the variability in subject head positioning in the scanner, the sparsity patterns appear to have different angles. In this experiment we evaluate our method by using high resolution MRI data which we downsample to simulate the sparse acquisition protocol. Specifically, we simulate a sparse scan for each subject by removing five out of every six slices at an arbitrarily rotated angle, then rotate the subject back to the common reference frame. We extract the middle coronal slice of each subject, perpendicular to the acquisition direction, and carry out 2D imputation for the middle coronal slice. We evaluate the methods described in the previous experiment. Because of the size of the images and dataset, we use a subspace encoding of 200 for the \textit{conv-FC} model. 

Figure~\ref{fig:graph_results} (right) summarizes reconstruction results across the entire test set, and Figure~\ref{fig:mri_imputation} illustrates example imputation of MRI slices. The spatial linear interpolation image introduces several artifacts. The reconstructions using the linear subspace is unable to reconstruct detailed anatomy. In contrast, the proposed variants achieve significantly better reconstruction, demonstrating the promise of using the proposed methods on clinical MR images.

\subsection{Evaluation with Deep Image Priors}

We evaluate our model, focusing on the \textit{conv-sparse} variant, compared to the deep image priors (DIP)~\cite{ulyanov2017deep}. We applied both methods to sparse data, as described in Section~\ref{sec:compare-discuss}. For a direct comparison, we use the same decoder architecture, described in our model variants, for both the proposed model and DIP. We use the MNIST-2 and FASHION-MNIST-1 datasets in various high sparsity scenarios, as well as the sparse brain MRI scans used in the previous section. Figure~\ref{fig:dip} summarizes the results. As the sparsity level increases (fewer pixels observed), the Deep Image Prior strategy lacks sufficient data in a single image to synthesize a reasonable image. In contrast, our method is able to leverage the entire dataset of sparse images with common structure to learn covariation within pixels, enabling better imputation of missing pixels. Our method requires 18.6 $\pm$ 13.7 \textit{ms} (since we only need to evaluate the network) to reconstruct an MRI brain slice. Deep Image Priors require learning network parameters, which took \mbox{36.5 $\pm$ 0.2} \textit{seconds} for images in our dataset\footnote{We report the runtime of Deep Image Prior for 1000 iterations, which we found to be necessary for good results}.



\section{Conclusion}

In this paper, we introduce a general probabilistic model to characterize sparse high dimensional imaging data by a deep non-linear subspace. We provide a principled derivation of the learning strategy in the presence of missing data, and introduce missing data aware network building blocks. We describe how missing data changes existing subspace models like the VAE, and how our model can be seen as a global version of deep image priors. 
Given a learned non-linear model, we describe how imputation can be achieved.

In our experiments, we demonstrate the importance of a novel sparsity-aware fully connected layer. We then show that our model exploits intra-image structure, as well as structure across a dataset, to yield superior imputation to fully convolutional architectures. Finally, we show the utility of our method using a real-world problem involving medical images.

{\small
	\bibliographystyle{ieee}
	\bibliography{references}
}

\end{document}